\documentclass[runningheads,a4paper]{llncs}

\usepackage{times}
\usepackage{amssymb}
\usepackage{graphicx}
\usepackage{url}
\usepackage[lined,plain,commentsnumbered,linesnumbered]{algorithm2e}

\urldef{\mailsa}\path|rochadan60@gmail.com,alex.lan@usp.br,ivandre@usp.br|

\begin{document}

\title{Semi-automatic definite description annotation:\\a first report}
\titlerunning{Semi-automatic definite description annotation}
\author{Danillo da Silva Rocha \and Alex Gwo Jen Lan \and  Ivandr\'e Paraboni}
\authorrunning{D.S.Rocha \and A.G.J. Lan \and  I.Paraboni}
\institute{University of S\~ao Paulo, Schools of Arts, Sciences and Humanites, S\~ao Paulo, Brazil 
\mailsa\\
}

\index{Rocha, Danillo}
\index{Lan, Alex}
\index{Paraboni, Ivandr\'e}

\toctitle{} \tocauthor{}

\maketitle

\begin{abstract}
Studies in Referring Expression Generation (REG) often make use of corpora of definite descriptions produced by human subjects in controlled experiments. Experiments of this kind, which are essential for the study of reference phenomena and many others, may however include a considerable amount of noise.  Human subjects may easily lack attention, or may simply misunderstand the task at hand and, as a result, the elicited data may include large proportions of ambiguous or ill-formed descriptions. In addition to that, REG corpora are usually collected for the study of semantics-related phenomena, and it is often the case that the elicited descriptions (and their input contexts) need to be annotated with their corresponding semantic properties. This, as in many other fields, may require considerable time and skilled annotators. As a means to tackle both kinds of difficulties - poor data quality and high annotation costs - this work discusses a semi-automatic method for the annotation of definite descriptions produced by human subjects in REG data collection experiments. The  method makes use of simple rules to establish associations between words and meanings, and is intended to facilitate  the design of experiments that produce REG corpora.
\end{abstract}

%%%%%%%%%%%%%%%%%%%
\section{Introduction}
%%%%%%%%%%%%%%%%%%%
\label{sec-intro}

Natural Language Generation (NLG) studies often make use of controlled experiments involving human subjects to elicit linguistic forms from (usually visual) stimuli. Experiments of this kind are motivated by the interest in knowing not only the text produced by the subjects, but also for exerting control over the stimuli (or context) that motivated the language production in the first place.

Of particular interest to the present work, we notice that REG experiments  - which addresses the computational problem of content selection - often make use of psycholinguistic methods to elicit corpora of definite descriptions. In these experiments, participants are instructed to provide a uniquely identifying description of a given target object. Examples of corpora developed for REG research include TUNA \cite{tuna-corpus}, GRE3D3/7 \cite{gre3d3,gre3d7}, Craft \cite{craft}, GenX \cite{genx}, Stars2 \cite{stars2}, ReferItGame \cite{Referit}, {\em b5-ref} \cite{b5-corpus,b5-ref} and others. 

As in many other research fields, REG experiments to collect human descriptions may include a considerable amount of noise. Subjects may easily lack attention, or may simply misunderstand the task at hand. As a result, the elicited data may include large proportions of ambiguous or ill-formed descriptions that, for the purpose of many REG studies, will have to be discarded. 

In addition to the issue of data quality, we notice that REG corpora are usually developed for the study of human strategies of reference production, for the implementation of computational REG models or for the study of semantics-related phenomena in general \cite{pechmann,time-course}. For that reason, it is often the case that both the elicited descriptions and their corresponding input contexts need to be annotated with their corresponding semantic properties and, as a result, the design of fully-annotated REG corpora may require considerable time and skilled annotators \cite{kappa}.

As a means two tackle both kinds of difficulties - poor data quality and high annotation costs - this work presents an initial experiment involving a semi-automatic method for definite description annotation produced by human subjects in REG data collection. The  method makes use of simple rules to establish associations between words and meanings, and is intended as a tool to aid the design of experiments that produce REG corpora.

%%%%%%%%%%%%%%%
\section{REG corpora}
%%%%%%%%%%%%%%%

Once the act of referring by means of a definite description has been decided \cite{esslli1999,reiter-book,book2002}, a REG algorithm \cite{greedy,incremental,graph} is invoked to determine its semantic contents. This generally involves selecting discriminatory properties \cite{olson}  provided that the use of redundant information does not lead to false conversational implicatures \cite{grice}. Although the standard Incremental approach \cite{incremental} remains popular in the field \cite{inlg2000,kelleher-algorithm,assessing,phd}, we notice that a number of corpus-based methods have been proposed \cite{bohnet2008,difabbrizio,viethenphd,thiago-svm,thiago-speaker-pref}.

% is text plus images
REG studies often require the construction of corpora of definite descriptions. Although REG corpora may in principle resemble `ordinary' text corpora that are ubiquitous in many NLP fields, resources of this kind are actually highly specialised. Unlike natural language understanding tasks in general - which are able to exploit a wide range of existing text corpora from the web or other sources - few NLG tasks can actually rely on text alone. For many NLG studies - including the case of REG - it is necessary to obtain not only the kinds of text to be generated, but also {\em the initial conditions} or {\em contexts} that motivated them in the first place. 

% involves subjects, validating and annotating
In order to identify possible links between input (context) and output (text), building a  REG corpus will usually require the design of controlled experiments involving human participants. In most cases, experiments of this kind provide textual or visual stimuli from which text or speech are elicited. After collection, pre-processing and annotation, the final product is a fully-annotated representation of the elicited descriptions and their contexts (which in most studies in REG consists of a set of images).

% annotation
The annotation of a REG corpus consists of labelling both descriptions and the objects of each context (e.g., image elements) with their semantic properties represented as attribute-value pairs. To this end, the domain semantics is carefully designed according to the objectives of the study that motivated the data collection. Once the scheme is defined, description annotation follows the same rules applied to the image annotation task. %For instance, a description as in `the blue chair' from the TUNA domain \cite{tuna-corpus} may be represented as a set of properties as in \{{\em type}-chair, {\em colour}-blue\}. 

% tuna
A prominent example of a corpus for REG is the TUNA \cite{tuna-corpus} corpus, an annotated  collection of definite descriptions developed for the study of reference phenomena and REG algorithms in general. TUNA contemplates situations of reference in two domains: furniture pieces, and people's photographs. TUNA descriptions were elicited in controlled experiments conducted with 60 native or fluent speakers of English, comprising 2280 atomic expressions (780 singular and 1500 plural references) and their corresponding contexts. As in other resources of this kind, TUNA descriptions and their contexts are accompanied by semantic annotation. TUNA was the first large-scale REG corpus to be made publicly available for research, and it has been widely reused as training and test data in a wide range of REG studies, including three REG shared tasks \cite{reg2007,reg2008,reg2009}.

% gre3d3
The issue of relational reference - absent from the TUNA corpus - was a possible motivation for two subsequent projects that became also widely reused in the field: the {GRE3D3} \cite{gre3d3} corpus and its extension {GRE3D7} \cite{gre3d7}, in both cases addressing the issue of spatial reference in simplified three-dimensional visual contexts. In these two experiments, participants were instructed to describe geometric objects (spheres, cubes etc.) as in `the ball next to the red cube'. Put together, {GRE3D3} and {GRE3D7} convey 5110 descriptions produced by 350 participants, and are possibly the largest datasets of this kind available for research purposes, and are among the few REG corpora to contemplate spatial referring expressions \cite{diego-space,give}.

%%%%%%%%%%%%%%%
\section{Current work}
%%%%%%%%%%%%%%%

Corpus-based REG studies often require the annotation of large collections of definite descriptions, and since results based on the elicited data depend fundamentally on the accuracy of the annotation, it may not be realistic to assume that this procedure could (or should) be fully automated. However, as in many NLP annotation tasks, the semantic annotation of definite descriptions for REG may in principle benefit from semi-automatic methods that provide initial (and possibly incomplete) annotation information to be revised by human specialists at a later stage. A method of this kind - which can be seen as a shallow language understanding task\footnote{Not to be mistaken for anaphora resolution \cite{msc1997,ramon,coling1998}.} - is the focus of the present work.

In REG studies such as TUNA \cite{tuna-corpus} or GRE3D3 / 7 \cite{gre3d3,gre3d7}, it is often the case that the elicited descriptions have a fairly simple syntactic structure and, as a result, we notice that a shallow parsing method (e.g., based on simple word-meaning associations) may be sufficient for the semi-automatic annotation of their semantic properties. For instance, a TUNA description as in `the red couch' may be easily interpreted as ({\em type}-couch, {\em colour}-red) even without full syntactic analysis.

Based on these observations, we envisage an annotation method that makes use of simple heuristic rules to establish associations between words string and the semantic properties that they represent. This method requires a knowledge base representing all possible word-property mappings for the relevant domain, but since a typical REG experiment would normally provide detailed domain semantics as part of its own design, building a knowledge base of this kind is unlikely to add much cost to the task. Table \ref{tab-mappings} illustrates a number of examples of word-property mappings in the TUNA and GRE3D3/7 domains.

\footnotesize{
\begin{table}[ht]
\centering
\caption{Examples of word-property mappings}
{\footnotesize
\begin{tabular}{  l c c }
\hline
Domain	&	Words & Properties \\
\hline
TUNA-Furniture		& \{large, big, larger\} & {\em size}-large	\\
TUNA-People					&  \{man, guy, person\} & {\em type}-person \\
GRE3D3										&	\{ball, sphere\} & {\em type-ball}\\
GRE3D7										& \{above, on top of\} & {\em above}-lm\\
\hline
\end{tabular}
}
\label{tab-mappings}
\end{table}
}
\normalsize

The proposed annotation method works as follows. Let $D$ be a REG domain (which is typically provided by the underlying REG project for which the data have been elicited) consisting of a set of objects and their possible properties represented as attribute-value pairs.  Given a list of words $S$ representing an elicited definite description for a target object $r$, and given a set of $M$ word-property mappings applicable to the domain $D$, the goal of the annotation method is to compute the set $L$ of all properties of $r$ that could be identified from the words in $S$. The following algorithm illustrates this procedure.

\footnotesize{
\begin{algorithm}[htp]
\label{alg:code}
\footnotesize{
	\DontPrintSemicolon
	\SetKwFunction{algo}{Heuristic}
	\SetKwProg{myalg}{}{}{}
	\myalg{\algo{$S$, $M$, $D$, $lang$}}{
    	\If {$lang==English$} {
					$Reverse(S) $\\
			}
		$L \leftarrow \emptyset$\\
  $Z \leftarrow Split(S,D)$\\
	\For{ $z_{i} \in Z$} {
			\For{ $w_{j} \in z_{i}$} {
								$np \leftarrow  NearestNoun(w_{j},  z_{i}$)\\
						    $p \leftarrow M[w_{j}+np]$\\
								\If {$p \neq null$} {
											$L \leftarrow L\cup p$\\
								}
								\Else{
									$p \leftarrow M[w_{j}]$\\
									\If {$p \neq null$} {
											$L \leftarrow L\cup p$\\
									}
								}
		
			}	 % For palavras wj

			\KwRet $L$\;
	} % For subcomponents si
	{}
} % algorithm
} % footnotesize
\caption{Shallow parsing}
\end{algorithm}
}
\normalsize

\begin{table*}[ht]
\centering
\caption{Test datasets}
{\footnotesize
\begin{tabular}{  l c c  l}
\hline
Domain	&	Training & Test 		&	Exemple \\
\hline
TUNA-Furniture		&  63								& 288	& the large red couch \\
TUNA-People		& 54												& 303 			&  the man with gray beard and glasses  \\
GRE3D3		&	90																	& 540 			& the small green cube \\
GRE3D7		& 624																& 3856 		& the red ball next to a large cube \\
\hline
\end{tabular}
}
\label{tab-corpus}
\end{table*}

The method is intended to handle descriptions produced in either English or Portuguese. However, since the English language places ordinary modifiers before the head noun (e.g., `red ball') and Portuguese places them after (e.g., `bola vermelha', or literally `ball red'), the algorithm starts by checking whether the language {\em lang} of the description is English (line 2). If so, the input string $S$ is reversed  (line 3) so that the head noun will be inspected first (as it would be the case in Portuguese). %In other words, English descriptions are reversed and subsequently treated in the same way as their equivalent in Portuguese.

After the language-specific treatment, an empty output set $L$ is created  (line 5), and the auxiliary function {\em in Split} (not detailed) is invoked to split the input $S$ In $k$ subcomponents $z_ {1..k}$ separated by relational properties (6). The purpose of this procedure is to allow each referent in a relational expression to be processed individually. For instance, the description `the green ball {\em near} a blue cube' refers to a main target object (the ball described by the left portion of the string), and to a landmark object (the cube described by the right portion of the string), and the boundary between the two is defined by the relational property `near'. Deciding whether a particular property is relational (e.g., {\em near}-lm) or not (e.g., {\em  colour}-red) follows from the domain definition $D$, since the only applicable values for relational properties are domain object identifiers (e.g., `lm'), and not ordinary object features (e.g., `red'). 

Once split, each {\em substring} $z_i$ is treated individually (line 7) and its words are considered in association with the nearest noun (9-13) or, if necessary, in isolation (14-20). At first, the nearest noun $np$ (which is the likely head noun to which $w_j$ is subordinate) is located by the auxiliary function {\em NearestNoun} (not detailed in the code) in the appropriate (English or Portuguese) language direction as defined at the beginning of the algorithm (9). Next, the combination of the current word $w_j$ and the head noun  $np$ (e.g., `black hair') is looked up in $M$ to verify whether it corresponds to a property $p$ (10) as in {\em hair.colour}-dark. If so (11), $p$ is added to output set $L$ (12). 

Despite the simplicity, this procedure allows the correct identification of most dependencies found in our test data (to be discussed in the next section) without the need for full syntactic analysis, and may be arguably considered sufficient for the present purposes. For example, it is possible to determine the correct meaning of `dark ' in expressions such as` dark man' and `man with dark beard ' as {\em hair.colour}-dark or {\em beard.colour}-dark, respectively. We notice however that this kind of dependency is relatively rare in our test corpora, and it is limited almost exclusively to the TUNA-People domain. For the vast majority of cases, the direct  match between individual words and properties is the most frequent outcome, that is, existing descriptions tend to present one-to-one mappings from words to properties. For example, in `large blue box ', each word corresponds to exactly one property in the GRE3D3 domain definition \cite{gre3d3}.

If $M$ does not contain any mappings from $w_j + np$ to a domain property, the word $w_j$ is searched individually, that is, without considering possible associations to any specific noun (15). If a mapping between $w_j$ and a property $p$ is found (16), $p$ is added to the output set $L$ (17). The procedure is repeated until all words of all {\em substrings} have been considered, and then the output set $L$ is returned (21). Throughout this process, unidentified words are disregarded, and the resulting $L$ may remain incomplete or even empty.

%%%%%%%%%%%%%%%
\section{Evaluation}
%%%%%%%%%%%%%%%

This section describes the evaluation of the proposed method based on four semantically-annotated definite description corpora. The objective of the evaluation is to measure the degree of proximity between the existing annotation and the one that would be obtained by using the proposed method.

%....................................
\smallskip
\noindent
{\bf Computational models} For the purpose of evaluating our Heuristic-based method, we consider as a baseline system a semi-automatic annotation alternative implemented using a neural POS-tagger called {\em nlpnet}\footnote{\url{http://nilc.icmc.usp.br/nlpnet/}.} that has been presently adapted to generate labels representing semantic properties rather than POS information.

The use of {\em nlpnet} as a tool for definite description annotation will be hereby referred to as the baseline POS method. This method requires training data represented as a set of manually labelled examples (which correspond to the 'semi-automatic' aspect of the method), and from the resulting model previously unseen test data may be labelled.

Both methods - Heuristic and POS - are logically equivalent in the sense that both make use of the same input knowledge, and are essentially distinguished only by the way in which this knowledge is represented (as mappings from words to semantic properties, or as labelled examples of definite descriptions). % The Heuristic method may however be more convenient for real-time use, as in the case of wanting to annotate (and possibly treat inconsistencies etc.) in the description already during the data collection experiment. The POS method, on the other hand, is best suited for batch processing of sets of descriptions previously collected.

%....................................
\smallskip
\noindent
{\bf Data} For the evaluation of the Heuristic and POS annotation methods, we consider the semantic annotation available from four REG domains: TUNA-Furniture and TUNA-People \cite{tuna-corpus}, GRE3D3 \cite{gre3d3} and GRE3D7 \cite{gre3d7}. For each of the four domains, a small portion of data (from 14\% to 18\%) was selected for training each model, and the remainder was reserved for testing. Table \ref{tab-corpus} shows the number of training and test instances (i.e., descriptions) and linguistic examples observed in each domain.

%....................................
\smallskip
\noindent
{\bf Procedure} In the case of the Heuristic method, the training set was taken as the source for extracting the mappings between words and their corresponding properties, as discussed in the previous section. In the case of the POS baseline method, the training dataset was manually annotated with labels representing the properties of interest, and taken as training data for a {\em nlpnet} tagger.  The actual evaluation consisted in applying the two models to the four test datasets, comparing their results with the reference annotation available from each corpus. As in much of the work in REG, we measured Dice \cite{dice} and Accuracy scores\footnote{This contrasts the use of surface realisation metrics such as BLEU \cite{bleu} and NIST \cite{nist}. }. Dice coefficients range from 0 to 1, in which 1 represents total coincidence between semi-automatic and corpus annotations. Accuracy represents the proportion of cases in which the two annotations were identical.

%....................................
\smallskip
\noindent
{\bf Results} Table \ref{tab-results} presents mean Dice and Accuracy scores obtained by the two methods under evaluation for each test corpus. Statistically significant differences between the two methods are highlighted.

\begin{table}[ht]
\centering
\caption{Results}
{\footnotesize
\begin{tabular}{  l c c c c }
\hline
\multicolumn{1}{c}{}&
\multicolumn{2}{c}{Heuristic}&
\multicolumn{2}{c}{POS}\\
Test corpus & Dice & Acc. 									& Dice & Acc.\\
\hline
TUNA-Furniture 	& {\bf 0.83} 						& {\bf 0.47} & 0.63 & 0.09 \\
TUNA-People 			& 0.38 & 0.01 				& {\bf 0.50} & {\bf 0.12} \\
GRE3D3 								& {\bf 0.99} 						& {\bf 0.96} & 0.76 & 0.74 \\
GRE3D7 								& {\bf 0.97} 						& 0.86 & 0.95 & 0.92 \\
\hline
\label{tab-results}
\end{tabular}
}
\end{table}

The Heuristic method generally outperforms the baseline POS method, except for the TUNA-People domain. High similarities with the corpus annotation - close to 100 \% - are observed in the GRE3D3 and GRE3D7 domains, whereas for TUNA-Furniture and, in particular, for TUNA-People, results are more modest. A possible reason for this outcome is the greater lexical variety (especially in the case of TUNA-People), which may not have been sufficiently represented in our small training dataset.

The comparison between Dice scores obtained by the two methods was performed using a Wilcoxon signed-rank test. Dice scores obtained by the Heuristic method are, on average, significantly higher than those obtained by POS for TUNA-Furniture ($W$=23027, $Z$=1.85, $p<$0.001), GRE3D3 ($W$=10072, $Z$=10.15, $p<$0.001) and GRE3D7 ($W$=17180, $Z$=3.03, $p=$0.0024) domains. In the TUNA-People domain, an opposite effect was observed ($W$=-14630, $Z$=-6.13, $p<$0.001). 

Finally, the comparison between accuracy values was performed using the Chi-square test. The accuracy observed by the Heuristic method is, on average, significantly higher than POS accuracy for TUNA-Furniture ($\chi^{2}=$ 35.81, df =1, p$<$0.01), and GRE3D3  ($\chi^{2}=$ 18.98, df =1, p$=$0.00013). In the case of the TUNA-People domain, an opposite effect was observed ($\chi^{2}=$ 9.95, df =1, p$=$0.001604), and in the case of GRE3D7, the difference between the two methods was not significant.

%%%%%%%%%%%%%%%%%%%%%
\section{Final remarks}
%%%%%%%%%%%%%%%%%%%%%

This work presented a first experiment using a semi-automatic method for definite description annotation. The method is based on simple heuristics that associate words to the most likely properties that they represent, and it is intended to simplify the annotation of large datasets that are typically elicited in REG studies. Moreover, since our approach does not require a large training dataset in the form of linguistic examples, it may be especially suitable for real-time validation of individual descriptions elicited during a data collection task.

As a future work, the current method will be embedded in an online tool for conducting REG experiments that provide feedback to the participants (e.g., about ambiguity and ill-formedness). This will enable participants to rephrase their answers if necessary and, as a result, may help the collection of higher-quality  data accompanied by tentative semantic annotation.

%%%%%%%%%%%%%%%%%%%%%%%%
\section{Acknowledgements}
%%%%%%%%%%%%%%%%%%%%%%%%
This work has been supported by grant \# 2016/14223-0, S\~ao Paulo Research Foundation (FAPESP).

%%%%%%%%%%%%%%%%%%%%%%%%
\section{References}
%%%%%%%%%%%%%%%%%%%%%%%%
\bibliographystyle{splncs}
\bibliography{refs}
\end{document}